%% file: main.tex
\documentclass[11pt]{article}

\usepackage{setspace,graphicx,epstopdf,amsmath,amsfonts,amssymb,amsthm}
\usepackage{marginnote,datetime,enumitem,subfigure,rotating,fancyvrb}
\usepackage{hyperref}
\usepackage{url}
\usepackage{float}
\usepackage[numbers]{natbib}
\usdate

\usepackage[left=2.5cm,right=2.5cm,bottom=2.5cm,top=2.5cm]{geometry}

\usepackage[flushleft]{threeparttable}
\usepackage[font=small,labelsep=period,justification=justified,singlelinecheck=false]{caption}
\usepackage{multirow}
\makeatletter 
\g@addto@macro\TPT@defaults{\footnotesize} 
\makeatother
\usepackage{array}
\usepackage{tabularx,ragged2e}
\newcolumntype{Y}{>{\RaggedRight\arraybackslash}X}
\newcolumntype{H}{>{\setbox0=\hbox\bgroup}c<{\egroup}@{}} 
\usepackage{pdflscape}
\newcommand{\smalltablefont}{\fontsize{9}{13.5}\selectfont}
\newcommand{\smalltablenotefont}{\fontsize{8}{12}\selectfont}

\newcommand{\mytilde}{\raise.17ex\hbox{$\scriptstyle\mathtt{\sim}$}}

\usepackage{booktabs}
\usepackage{my_style}
\usepackage{sectsty}

\sectionfont{\centering\uppercase}
\subsectionfont{\textbf}
\subsubsectionfont{\normalfont\textit}

\title{Discovering Significant Topics from Legal Decisions with Selective Inference}
\author{\uppercase{Jerrold \textbf{Soh} Tsin Howe}\thanks{Yong Pung How School of Law, Singapore Management University (jerroldsoh@smu.edu.sg, https://orcid.org/0000-0003-0270-5015). This is an accepted manuscript of work forthcoming in PhilTrans A; please cite the publisher's version only. I thank Georgios Georgiou, James Greiner, Alma Cohen, Lucian Bebchuk, William Hubbard, Michael Livermore, and Kevin Ashley, and participants of the Complexity in Law \& Governance Works in Progress Workshop for constructive comments, including on very early versions of this work. All errors are mine.}}

\begin{document}
\maketitle
\begin{abstract}

    \noindent We propose and evaluate an automated pipeline for discovering significant topics from legal decision texts by passing features synthesized with topic models through penalised regressions and post-selection significance tests. The method identifies case topics significantly correlated with outcomes, topic-word distributions which can be manually-interpreted to gain insights about significant topics, and case-topic weights which can be used to identify representative cases for each topic. We demonstrate the method on a new dataset of domain name disputes and a canonical dataset of European Court of Human Rights violation cases. Topic models based on latent semantic analysis as well as language model embeddings are evaluated. We show that topics derived by the pipeline are consistent with legal doctrines in both areas and can be useful in other related legal analysis tasks.

\end{abstract}

\setlist{noitemsep}  

\newpage
\singlespacing

\newpage

\pagestyle{myheadings}
\pagenumbering{arabic}

\section{Introduction}
\thispagestyle{plain}
\label{sec:introduction}

Most legal information is stored exclusively in natural language texts. The complexity of language means extracting such information is typically a labour-intensive exercise primarily performed by specially-trained persons (``lawyers''). This poses significant barriers to computational representations and analysis of law \cite{mccarty_deep_2007,nazarenko_pragmatic_2021}. Researchers have increasingly sought to develop automated processes for converting unstructured legal texts to structured variables \cite{grimmer_stewart_2013, alscher_et_al_2017}. Depending on the texts involved and variables required, these have included term frequency counts \cite{choi_2020}, regular expressions \cite{soh_2019_network}, topic models \cite{aletras_2016,falakmasir_2017,carter_brown_rahmani_2018,dyevre_glavina_ovádek_2021, salaun_why_2022}, word embeddings \cite{dhanani_et_al_2022, jayasinghe-etal-2022-learning}, and language models \cite{chalkidis-etal-2020-legal, shounak_et_al_2023, licari_et_al_2023}. Given their centrality in legal analysis, court decisions in particular have attracted significant scholarly attention. Many studies have attempted to identify, categorise, or forecast case outcomes using decision texts, often relying on opaque algorithms such as support vector machines and neural networks \cite{aletras_2016, liu_chen_2017, chalkidis-etal-2019-neural, medvedeva_et_al_2020, medvedeva_et_al_2023}. Other researchers have prioritised more explainable methods over end-to-end prediction. Typically, algorithms are first developed to automatically extract case attributes and other legally-relevant variables before using these variables to model outcomes \cite{ashley_bruninghaus_2009, falakmasir_2017, branting_et_al_2021, Gray2022, gray_et_al_2023_automatic}. The goal is not necessarily predictive accuracy alone, but also to identify and explain what motivates legal decisions.

In this work, we propose and evaluate a new automated pipeline for discovering significant topics from decision texts, a task we define more formally in part \ref{meth:defn}. The pipeline takes decision texts and case outcomes as inputs and returns estimates for statistically significant decision topics as well as the cases, words, and phrases most strongly associated those topics. This allows researchers to quickly identify potential variables, patterns, and cases of interest in unfamiliar areas of law. The pipeline comprises four steps: pre-processing and masking (part \ref{meth:preproc}), topic modelling (part \ref{meth:topic}), selective regression and inference (part \ref{meth:lasso}), and topic evaluation (part \ref{meth:eval}). We demonstrate and evaluate the pipeline on a new dataset of cases resolved under the Uniform Domain Name Dispute Resolution Policy (``UDRP''). To explore how the pipeline generalises, we further test it on a canonical dataset of European Convention of Human Rights (``ECHR'') cases. For both datasets, we experiment with Latent Semantic Analysis (``LSA'') \cite{landauer_1998} as well as two BERTopic (``BTO'') models \cite{grootendorst2022bertopic} primed with general and legally-finetuned embeddings respectively. 

We show that topics discovered by the pipeline contain interpretable and legally-sound information on topics correlated with legal outcomes (part \ref{sec:results}). Along the way, we identify several interesting patterns and case archetypes in UDRP and ECHR case law. Thus, our key contributions are as follows. First, we extend prior work analyzing legal outcomes from a topic modelling perspective \cite{aletras_2016, falakmasir_2017, salaun_why_2022}. To be sure, the notion that topics synthesized from case decisions could carry meaningful information about legal outcomes is not new. Neither do we propose entirely new algorithms for, say, legal topic modelling. Our incremental contribution lies in integrating several existing techniques (e.g.\ masking \cite{sulea_zampieri_vela_genabith_2017}, topic modelling \cite{aletras_2016, salaun_why_2022}, and selective inference \cite{Tibshirani94regressionshrinkage, taylor_tibshirani_2017}) into a pipeline that can be adapted to study other legal areas. Second, we demonstrate the utility of selective inference techniques in the legal domain. This has not, to our best knowledge, been studied in prior work. Finally, we add to legal knowledge on UDRP and ECHR cases.

\section{Methods}
\label{sec:methods}

\subsection{Discovering Significant Topics}
\label{meth:defn}

This work relates to existing literature on the automated extraction of legal factors from legal cases \cite{falakmasir_2017, Gray2022, gray_et_al_2023_automatic}. Legal factors are generally seen as ``stereotypical patterns of fact'' \cite{falakmasir_2017} or more abstract ``intermediate concepts'' \cite{canavotto_horty_2023} which influence case outcomes. However, as used in that literature, the concept of legal factors has a specific meaning which does not overlap perfectly with our present focus. We thus use the term ``predictors'' here to refer broadly to variables which predict case outcomes. Drawing inspiration from \cite{chen_2018}, suppose a legal outcome $Y$ is given by $Y=f(X, W)$, where $X$ is a matrix of legal predictors, $W$ a matrix of non-legal predictors (e.g.\ political ideologies \cite{ruger_kim_martin_quinn_2004}), and $f$ some adjudication function that maps cases to outcomes. To identify individual predictors, we might collect data on hypothesised variables $\hat{X}, \hat{W}$ (i.e.\ approximations of $X$ and $W$), and estimate the model $\hat{Y} = \hat{f}(\hat{X}, \hat{W})$. Weights computed for each $\hat{x}, \hat{w}$ would capture the strength and polarity of their correlation with outcomes. Variables assigned significant, non-zero weights can be understood as potential legal (or non-legal) predictors. They may further be seen as \textit{causal} predictors, if the model is causally-identified, or correlative predictors otherwise.

The challenge with legal applications is that the $x$s and $w$s are not available as structured data but found only in some natural language corpus $D$. Typically these are decision texts written to state and justify outcomes for each case $i$, though other documents including submissions, affidavits, and procedural records may also be relevant. We must apply a ``codebook function'' $g:D\mapsto Q, Q \in \mathbb{R}^{n \times m}$ that maps $n$ texts to $m$ variables \citep{grimmer_stewart_2013}. Where the variables desired are known \textit{ex ante} based on legal domain knowledge, the researcher's aim is to extract \textit{observations} of $\hat{x}_{i}$. But in unfamiliar legal areas where candidate predictors are not already known, the goal shifts from filling observations or estimating coefficients to \textit{discovering} such predictors to begin with. There are therefore three different tasks related to legal predictors (table \ref{tab:task_names}). Notably, these are not mutually exclusive and must often be performed in tandem to answer the research question. Suppose as in \cite{soh2021Causal} that we want to know if case origin influences the probability of a certiorari grant by the US Supreme Court. The variable of interest is known but potential confounders remain to be identified. We would need to extract observations for case origin, discover (and thereafter extract observations for) potential confounders, and finally analyse coefficients for case origin while controlling for these confounders. This work is chiefly concerned with the discovery task, though extraction and analysis are by-products of the proposed method.

\input{tables/task_table}

\subsection{Proposed Pipeline}

\subsubsection{Pre-processing and Masking}
\label{meth:preproc}

We begin with a text corpus $D$ and structured categorical outcomes $Y$ for $n$ cases in some legal area of interest. In theory, any corpus with sufficient case information, such as case briefs and affidavits, could be used. In practice, most legal analysis is based solely on decision texts. Other legal documents are usually not accessible at scale. Thus, we tailor the approach assuming $D$ is a decision corpus. The use of decision texts has important implications for the kind of analyses possible and the pre-processing steps necessary. Specifically, fitting legal outcome models on decisions is problematic because decisions are written by judges, after observing case facts, to justify case outcomes \cite{soh2021Causal}. Extracted features could therefore contain both post-treatment and post-outcome information, making them ``bad controls'' \cite{cinelli2022}. Formally, suppose decisions are generated by the process $D = t(Y, X, W, J)$, where $J$ accounts for the judges' individual writing styles and $t$ some text-generation function. Substituting this into the model $\hat{Y} = g(D)$ gives $\hat{Y} = g(t(Y, X, W, J))$. Since we are indirectly modelling $Y$ on itself, we should expect the model to produce large, significant estimates for features still containing hints of $Y$ after the transformations $g$ and $t$ instead of unbiased estimates for $x$s and $w$s.

As we do not control $t$, the natural solution, other than switching to some pre-outcome corpus, is to build into $g$ processes for masking information on $Y$. We follow standard steps from the legal prediction literature in masking outcome-revealing sections of and phrases in the text from the model by deleting them entirely at the start of the pipeline \cite{aletras_2016, sulea_zampieri_vela_genabith_2017}. This may over-inclusively remove otherwise informative words, but is however taken as a necessary and non-fatal trade-off \cite{sulea_zampieri_vela_genabith_2017}. It also may not remove all outcome information from $D$. Since decisions are written to \textit{justify} case outcomes, even seemingly innocuous sections such as ``Case Facts'' could be arranged in a way that favours the writer's preferred disposition. Indeed, lawyers are typically taught to present facts \textit{persuasively} \cite{georgetown_law_guide}. This pertains especially to case briefs, but we cannot preclude its occurrence in decisions. As such, we emphasise that predictors discovered by our method should be interpreted as \textit{correlative}.

Where required, we then pre-process the masked corpus in standard fashion by lowercasing, stopping, and lemmatisation. This applies mainly to LSA as BTO is trained on raw texts.\footnote{\url{https://maartengr.github.io/BERTopic/faq.html\#how-do-i-remove-stop-words}}

\subsubsection{Topic Modelling}
\label{meth:topic}

Topic models are suitable codebooks because of their readability: each $q \in Q$ can be manually interpreted based on representative n-grams, and documents with higher $q$ weights can be read as being more heavily or likely `about' $q$. Of the numerous topic models in the literature, here we experiment with one hot encoding (i.e.\ indicators for each n-gram in the corpus' overall vocabulary) (``OHE''), LSA, and BTO to cover a range of traditional and emerging approaches. As topic models are well-documented elsewhere, below we provide a condensed description of those we test. 

LSA first computes a term-frequency/inverse-document-frequency (``TFIDF'') encoding \citep{jones_1972, salton_buckley_1988, dominich_2001}. The TFIDF matrix is compressed into $m$ desired topics (explained below) by applying Singular Value Decomposition (``SVD'') and keeping only features corresponding to the largest $m$ singular values. The SVD of a matrix $W = U_mS_m\Lambda_m^T$ where $rank(W) = m, m \leqslant rank(W)$ \cite{bishop_2006}. When $W$ is a TF-IDF matrix, $U_m$ corresponds to n-gram vectors, $S_m$ to singular values of $W$, and $\Lambda_m$ to document vectors \citep{dominich_2001}. The corpus is thus represented through $\Lambda_m$ as a distribution of $m$ topics across $n$ documents \citep{landauer_1998}. These ``topics'' are represented in $U_m$ as distributions across n-grams. For intuition, observe that an optimal compression of term frequency matrices should squeeze co-informative terms together, forming said topics. We use LSA here because of its prominence in \cite{aletras_2016}'s influential work on legal outcome prediction for ECHR cases as well as subsequent related work.

BTO \cite{grootendorst2022bertopic} is modular framework which starts with paragraph embeddings typically derived from a language model. Depending on the LM's context window, longer documents may be partitioned into smaller chunks if necessary \cite{silveira_et_al_2021}. Chunk embeddings undergo dimensionality reduction via a standard algorithm such as UMAP \cite{mcInnes2018} (the default) or principal components analysis before clustering via another algorithm such as HDBSCAN \cite{mcInnes2017} or k-Means. Topics are extracted from these clusters using a bag-of-words vectorizer followed by a ``class-based'' TFIDF implementation given by $cTFIDF(c) = ||tf_{w,c}|| \times log(1+\frac{A}{f_w})$ where $tf_{w,c}$ is the frequency of n-gram $x$ in cluster $c$, $f_w$ is $w$'s' frequency across all clusters, and $A$ is the average number of tokens per cluster. This produces an arbitrary number of topics which can be iteratively merged based on topic frequency and $cTFIDF$ similarity until a desired number remains. The resulting chunk-topic matrix can then be re-constituted into document-level topics in several ways. For instance, by assigning a document to the one topic which contains the largest number of its chunks (i.e.\ max-pooling). Following \cite{silveira_et_al_2021}, we take chunk-topic counts normalised at document level. We test two BTO models primed with chunk embeddings from (1) all-MiniLM-L6-v2 \cite{minilm_site}, a sentence transformer based on \cite{minilm} and recommended by \cite{grootendorst2022bertopic} (``BTO\textsubscript{M}''), and (2) legalBERT \cite{chalkidis-etal-2020-legal}, a BERT \cite{devlin-etal-2019-bert} extension fine-tuned on UK, EU, and US legal documents (``BTO\textsubscript{L}''). Inspiration for using BTO in the legal context comes from \cite{salaun_why_2022} which used a multilingual MiniLM-embedded BTO model to study Canadian housing law court decisions written in French.

Here we generate topics comprising $1,2,3$-grams for all topic models. For LSA, we generally take only the 2500 most frequent n-grams at the TFIDF step before reducing the matrix to a desired topic number based on corpus size. As context, \cite{aletras_2016}'s best predictive models for ECHR cases generally used LSA topics creating with the 2000 top $1,2,3,4$-grams. However, in our (unreported) exploratory tests, we noted that 4-grams do not add new interpretable information as they usually repeated terms already seen in $1,2,3$-grams. We also set minimum document frequency cutoffs of 5 or 10 (depending on dataset and topic model) in LSA's TFIDF and BTO's cTFIDF steps to limit computational and memory overheads. Other parameters follow recommendations and defaults from the \texttt{sklearn} \cite{sklearn_lsa} and \texttt{bertopic} \cite{grootendorst2022bertopic} libraries.

\subsubsection{LASSO Regression and Selective Inference}
\label{meth:lasso}

We use a LASSO \citep{Tibshirani94regressionshrinkage} regression model to associate topics with outcomes. The LASSO uses the coefficient vector's L1-norm as a penalty term when optimising the model, such that the objective function becomes $L(\beta)^* = L(\beta) - \lambda\lVert \beta_j \rVert$ where $\lambda$ is a user-specified ``shrinkage parameter'' that controls penalisation magnitude, and $j>0$ (the intercept is not penalised). The LASSO is suitable for legal outcome models in three ways. First, as the goal is to discover interpretable legal topics rather than inexplicably predict legal outcomes, regression models are preferable to more opaque approaches like neural networks. Second, the LASSO overcomes two common, related problems with legal outcome models. First, as text feature matrices are typically large and sparse, and legal corpora often yield few observations, legal outcome models are prone to the $k\gg{}n$ problem \citep{zou_hastie_2003, hastie_tibshirani_friedman_2017}:\ as $k$ approaches and eventually exceeds $n$ standard regression models relying on maximum likelihood estimation are liable to produce biased estimates or failing to converge entirely. Second, legal areas often present highly imbalanced response classes, forcing us to estimate ``rare events'' \citep{zorn_2005, bielza_robles_larranaga_2011, pavlou_2015}. For instance, in our UDRP dataset, \mytilde90 percent of the cases are decided in the complainant's favour (table \ref{tab:udrp_stats}). Coupled with $k\gg{}n$, legal outcome models could be perfectly separated --- outcomes can be perfectly predicted with a subset of features --- preventing model convergence. 

Penalised regressions are one standard countermeasure to both problems \citep{firth_1993, heinze_1999, zorn_2005, hastie_tibshirani_wainwright_2015, hastie_tibshirani_friedman_2017}. In bioinformatics and chemometrics, LASSO regressions have been successfully deployed in studies involving large feature matrices and rare events \citep{wu_chen_hastie_sobel_lange_2009, zhu_tan_cheang_2017}. Third, the LASSO lets us exploit emerging methods for selective inference. Conventionally, significance tests are not done with penalised regressions since regularisation means estimates are biased toward zero and not consistent \citep{hastie_tibshirani_friedman_2017}. Nonetheless, LASSO regressions were demonstrably capable of selecting the most significant regressors, particularly in a $k\gg{}n$ setting (see \cite{hastie_tibshirani_friedman_2017}). More recently, \cite{lee_sun_sun_taylor_2016} devised a method for conducting valid post-selection significance tests which \cite{taylor_tibshirani_2017} extend to the LASSO. P-values are computed after de-biasing the model post-selection \cite{lee_sun_sun_taylor_2016, taylor_tibshirani_2017}. Coefficient estimates must still be interpreted in light of the penalty, but p-values and standard errors remain valid and have been shown to be more reliable than non-adjusted values from subset-selected models \citep{lee_sun_sun_taylor_2016,hastie_tibshirani_friedman_2017}. Notably, if as cautioned above we confine ourselves to discovery \textit{correlative} rather than causal predictors, significance test validity is less of a concern. We use \cite{taylor_tibshirani_2017}'s R package \texttt{selectiveInference} \citep{selinf_R} and following their documentation estimate the LASSO with \texttt{glmnet} \citep{glmnet_R}.

\subsubsection{Evaluation}
\label{meth:eval}

We test several model specifications for the primary UDRP dataset, varying whether topic features are included and the topic model used (see part \ref{sec:data}). Each specification is also evaluated on standard measures of fit including the area under the receiver operating characteristic curve (``AUROC'') and the median deviance ratio (``MDR''). The latter summarises all deviance ratios reported by \texttt{glmnet} along the $\lambda$ fitting path and can be interpreted as the pseudo-$R^2$ \citep{glmnet_R}. We manually evaluate selected specifications by delving into topics with the largest positive or negative coefficients and the smallest p-values for those specifications. The author, who is legally-trained, then studied the topics' n-gram distributions and the cases most strongly associated with them to see how far they corresponded with topics known to be significant in legal doctrine. Notice that even if they do not, topics discovered this way could point to some yet unknown X or W driving legal outcomes. This step should therefore be informed by legal theory. To be sure, we do not suggest it can be fully-automated, nor that the method is sufficient to identify all legally-significant topics.

Other than evaluation, the method requires structured data in only two respects. First, labelled case outcomes are needed. While not considered in this work, existing methods for automated legal outcome extraction (e.g.\ \cite{aletras_2016, katz_bommarito_blackman_2017, chalkidis-etal-2019-neural, medvedeva_et_al_2023}) could be incorporated at an earlier pipeline step. Second, tailored pre-processing work is necessary to sectionise documents and to mask outcome-leaking information. Other than in these two areas, topics correlated with legal outcomes are automatically synthesized from the corpus, selected by the LASSO, and surfaced by post-selection significance tests. Prior domain knowledge of potential legal predictors within the given legal area is neither assumed nor required, though it would certainly be a bonus. Likewise, while structured case metadata is not strictly needed, any available variables can easily be included as additional covariates at the regression stage.

\subsection{Datasets}
\label{sec:data}

\subsubsection{Domain Name Disputes}
\label{sec:data:udrp}

The UDRP is a mandatory policy instituted in 1999 by the Internet Corporation for Assigned Names and Numbers (``ICANN'') for resolving disputes over generic top-level domains (``GTLD''s). Several countries have adopted similar policies for their country-coded top-level domains (``CCTLDs'') \citep{chik_2007}. Disputes are administered by ICANN-appointed Dispute Resolution Providers (``DRP''s). The largest DRP by disputes resolved is the World Intellectual Property Organisation (``WIPO''). Under the \textit{Rules for Uniform Domain Name Dispute Resolution Policy}, a case begins when a trademark holder files a complaint with a DRP. The DRP will ask the respondent for a written response, and thereafter assemble an adjudication panel of 1 or 3 panellists, depending on the parties' preferences. Under UDRP Article 4a, the complainant must show that (1) the contested domain is ``identical or confusingly similar'' to the complainant's trade or service mark; (2) the respondent does not have any ``rights or legitimate interests'' in the contested domain; \emph{and} the contested domain was ``registered and used in bad faith''. While parties may be represented by lawyers, all procedures are written and there are no physical hearings. If a complaint succeeds, the panel may order the domain to be transferred to the complainant or be cancelled altogether. Decisions are communicated to and enforced by the relevant domain name registrar \citep{mueller_root_2002}.

We obtained from WIPO's online database\footnote{\url{http://www.wipo.int/amc/en/domains/search/}} decision texts for WIPO-administered UDRP disputes decided on and between 1999 and 2016. Regular expressions were developed, by iterative testing on randomly-sampled decisions, to partition the texts into 8 archetypal sections. Case outcomes are typically stated in a final section titled ``The Decision'', and occasionally in a preceding section generally titled ``Discussion and Findings''. The latter details the panel's legal reasoning and analysis. Both sections were masked. Outcome labels ``transfer'', ``cancel'', and ``deny'' and linguistic variants thereof were also removed. This left only sections on case facts, parties involved, procedural history, and arguments presented for downstream processing. Decisions where fewer than all 8 sections could be detected, either because they were not in English or because of exceptional or missing headers, were excluded. This reduced the initially downloaded set of 27,634 raw cases into 22,653 usable observations.\footnote{Most of the 4,981 cases dropped here were non-English cases. This is consistent with official WIPO statistics, available at https://www.wipo.int/amc/en/domains/statistics, which show that around 85\%--88\% of cases each year are in English.}

Labelled outcomes and other structured variables were extracted from case summary tables on WIPO's website. Each table contains case number, decision date, the domains, parties, and panellists involved, and outcome. While only three outcomes (i.e.\ transfer, cancellation, or complaint denied) are possible per domain, cases with multiple domains could present mixtures (e.g.\ ``Complaint denied, transfer in part with dissenting opinion''). Nonetheless, the vast majority (98.87\%) of cases involved singular outcomes. By studying the data we found that outcome statements start with the outcome assigned to a majority of the contested domains (i.e.\ in the example above most domains would \textit{not} have been transferred). We thus binarised outcomes by recording 1 when the outcome statement begins with ``Transfer'' or ``Cancellation'', and 0 when it begins with ``Complaint denied''. Basic string methods were used to extract other variables from the tables, including the number of panellists, complainants, respondents, and domain names involved, whether the case involved GTLDs or CCTLDs, and year and month indicators. We also created indicators for repeat complainants (respondents) appearing in \textgreater100 (30) cases.

Identity indicators were also created for all panellists. We use this to demonstrate how the method could be instrumental for studying how judge identity influences legal outcomes, a staple in ``judicial behaviour'' research \cite{Epstein2021}. Legal scholars have debated the UDRP's merits \cite{brannigan_2004}, with critics alleging structural pro-complainant biases in the UDRP procedural rules \cite{mueller_2001, geist_2001, mueller_2002success, kelley_2002}. Proponents \cite{donahey_2001, ned_2002, kur_2002} countered that critics fail to account for specific case attributes. Empirical analyses have offered different explanations for high complainant success rates. \cite{kesan_gallo_2005} argued that case resolution efficiency was as important as apparent bias in determining provider choice, while \cite{klerman_2016}'s used an alternative linear regression methodology on \cite{kesan_gallo_2005}'s dataset of 2000--2001 cases.

\input{tables/udrp_stats}

Table \ref{tab:udrp_stats} summarises the dataset. It contains information on significantly more cases and variables than an earlier UDRP corpus compiled by \cite{branting_et_al_2021}. On this data we run the penalised logit regression:\ $$complainantwon_i = panelistidentity_i + panelsize_i + textfeatures_i + controls_i + \epsilon_i$$ where $complainantwon_i$ is a indicator for complaint success, $panelistidentity_i$ an indicator matrix for panellist involvement, $panelsize_i$ indicates if the case involved three panellists or one, $textfeatures_i$ is either an OHE, LSA, or BTO document-topic matrix. $controls_i$ are indicators for year and month, repeat player involvement, and whether the case involved GTLDs or CCTLDs. As indirect controls for dispute complexity, we also included the raw and processed word counts of the relevant decision, as well as the number of complainants, respondents, and domain names involved.

To investigate the topic models' impact, we estimate regressions with/without topic features across three settings: (A) only 1-panellist GTLD cases, (B) all GTLD cases, and (C) all cases. We partition the data by panel size and domain type because these give rise to qualitatively different case types. To evaluate models in the same regression setting on similar bases, we extract exactly 250 topics with each topic model. We chose 250 after some iterative testing with LSA because it represented a 90\% compression of the original TFIDF matrix (recall that the top 2,500 n-grams were used) but, as computed by the SVD, explained about 61\% of the variance in the same. Around the 250 mark, reducing (increasing) the number of topics led to more (less) than proportionate losses (gains) in variance explained. We used LSA rather than BTO models to experiment with topic number because re-estimating BTO models requires significantly more compute. There is some inevitable arbitrariness here as identifying the appropriate number of topics is a known challenge in topic modelling \cite{greene_how_2014}. Future work could study how emerging techniques for doing so (e.g.\ \cite{gerlach_network_2018, sbalchiero_topic_2020}) could be incorporated into our method.

All topic models are trained using only decisions within the relevant partition. This except for BTO chunk \textit{embeddings} (only the first step) which are pre-computed only once on the entire corpus and used across all settings, as the embedding process is computationally expensive. We also pre-computed the shrinkage parameter $\lambda$ to be used using specifications \emph{without} text features following the guideline suggested in \cite{negahban_2009,aletras_2016} to set $\lambda = 2\mathbf{E}[\lVert X^T\epsilon \rVert_\infty]$ where $\epsilon \sim N(0, \hat{\sigma}^2)$ and $\hat{\sigma}^2$ is the residual sum of squares from a simple linear regression of $y$ on all regressors. The same $\lambda$s were then used for mirror specifications \emph{with} text features. As a further baseline, we also tested specifications with white noise placebos \citep{mcshane2011}.

\subsubsection{European Convention on Human Rights violations}
\label{sec:data:echr}

The ECHR establishes fundamental human rights for signatory jurisdictions, including the prohibition of torture (Article 3), right to a fair trial (Article 6), and right to respect for private and family life (Article 8). The European Court of Human Rights (``ECtHR'') adjudicates complaints. The court publishes decision texts and ``case detail'' tables on its ``HUDOC'' database.\footnote{\url{http://hudoc.echr.coe.int/}} ECHR cases have been studied in several prior works \cite{aletras_2016,medvedeva_et_al_2020} and included in the benchmark LexGLUE \cite{chalkidis-etal-2022-lexglue}. While LexGLUE provides a large number of processed ECHR texts and outcomes, that dataset is not linked to case identifiers, making topic interpretation challenging. Here we use \cite{medvedeva_et_al_2020}'s dataset and replicate their pre-processing steps with their published code. We limit the analysis to training set cases with clear violation/non-violation outcomes (i.e.\ not filed in the dataset as ``both''). Below we focus on Articles 3, 6, and 8 which have the largest number of cases in this dataset. Following \cite{medvedeva_et_al_2020}, we use only text from the Procedure, Circumstances, and Relevant Law sections. Table \ref{tab:echr_stats} summarises the dataset. As our aim was to demonstrate generalisability, unlike with the UDRP we did not further extract new case variables. The main specification tested is $violation_i = textfeatures_i + \epsilon_i$ with $textfeatures_i$ being 100 topics synthesised using the above topic models.

\input{tables/echr_stats}

\section{Results}
\label{sec:results}

\subsection{UDRP results}

Table \ref{tab:udrp_res} summarises our primary results on the UDRP dataset. Columns 1--3 report baseline estimates computed without any text features for three main regression settings. Around 50 panellists are significant at $\alpha=0.05$ across these baselines even with several controls included (column 3), suggesting an association between their involvement and complaint outcomes. The association is notably weaker in the corresponding topic regressions with OHE, LSA, BTO\textsubscript{M}, and BTO\textsubscript{L} features added (columns 4, 5--7, 8--10, and 11--13). The topic regressions consistently yield fewer significant panellists, smaller panellist effects, and better model fits. Statistical significance can be observed shifting towards the topics instead. This can already be observed with simple OHE, but is clearest with the LSA regressions, where few panelists remain significant (9, 13, 8 in columns 5--7 versus 53, 50, 49 in columns 1--3). Across all regression settings, LSA consistently produces the largest number of significant topics and the highest fit scores. Column \ref{tab:udrp_res}.7 in particular yields 32 significant topics but only 8 significant panellists at $\alpha=0.05$ and the highest MDR (0.431) and AUROC (0.914). BTO\textsubscript{L} and BTO\textsubscript{M} yield more significant panellists and fewer significant topics, but are nonetheless superior to the non-text and white noise (column \ref{tab:udrp_res}.14) baselines, suggesting that these topic models also capture information on case features. The legally-finetuned BTO\textsubscript{L} performs slightly better than BTO\textsubscript{M} (MDR=0.295, AUROC=0.849 in column 13 in versus MDR=0.275, AUROC=0.838 in column 10), suggesting that domain adaptation helps.

\input{tables/udrp_res}

These results are relevant to legal debates on whether UDRP processes exhibit pro-complainant bias. While our correlative models \textit{cannot} establish the absence of bias, our findings are consistent with \cite{ned_2002, kur_2002}'s argument that high complaint success rates are better explained by case facts than structural pro-complainant biases. More importantly, our results suggest that the pipeline can automatically discover \textit{correlative} legal predictors from decision texts. This becomes clearer when inspecting the discovered topics. The 5 LSA, BTO\textsubscript{M}, and BTO\textsubscript{L} topics with smallest p-values in columns 7, 10, and 13 respectively are presented in table \ref{tab:udrp_words}). Some topics are intuitive. For example, the negative effect associated with LSA 17, a topic populated by n-gram variations on ``administratively deficient'', suggests logically that ``administratively deficient'' complaints correlate to worse complainant outcomes. Manual evaluation revealed that cases with the strongest weights for this topic indeed involved deficient complaints.\footnote{WIPO Case Nos.\ D2009-0021, D2011-1484, D2014-2277, D2014-1901, and D2016-0102.} 3 of these complaints were denied. Likewise, the top LSA 3 cases involved situations where the complainant provided incorrect ``contact information'' for the domain registrant and was asked to amend the complaint accordingly.\footnote{See WIPO Case Nos.\ D2010-0593, D2012-1002, D2012-1761, D2014-1333, and D2016-0341.}

\input{tables/udrp_words}

Other topics are less readable, but their underlying logic can be identified on closer inspection. For instance, LSA 19 and BTO\textsubscript{M} 5 are populated by references to famous trademarks and brands. These topics feature most strongly in complaints filed by large corporations which owned these and other famous marks, and typically against individuals who had registered variations on their brand names.\footnote{For LSA see WIPO Case Nos.\ D2009-1392, D2013-1265, D2011-0391, D2010-1878, D2015-1445. For BTO\textsubscript{M} see WIPO Case Nos.\ D2002-0760, D2006-0297, D2008-0416, D2011-0022, D2015-1936.}. For example, 3 of the top 5 LSA 19 cases involved the ``lego'' company suing for domains such as ``legosets101.com'' and ``legowolds.com''. Panels typically found evidence of bad faith in how respondents could not have registered these domains without knowing of the complainants' well-known marks. 9 of the top 10 complaints succeeded. The exception was a complaint filed by ``Hugo Boss'' for ``boss-watch.com'' and ``boss-world.com''. This was denied because the respondent had been selling watches under the ``BOSS'' mark in Hong Kong since the 1970s, before the complainant's mark was established.\footnote{WIPO Case No.\ D2015-1936}.

Consider also BTO\textsubscript{L} 69, which associates references to ``reverse domain name hijacking'' (``RDNH'') with lower complaint success rates. UDRP Rule 1 defines RDNH as ``using the Policy in bad faith to attempt to deprive a registered domain name holder of a domain name''. When RDNH is found, the complaint fails. Recall however that the masked texts used in topic modelling exclude the ``Discussion and Findings'' and ``Decision'' sections, so the model should not have information on whether RDNH occurred. Inspecting the cases here reveals that RDNH n-grams feature strongly in the included ``Contentions'' section when respondents actively defend the claim and raise the RDNH issue. In the usual case where respondents default, neither panellists nor complainants have incentives or need to discuss it. Thus while RDNH was not ultimately found in any of the top 5 BTO\textsubscript{L} 69 cases, all were rare cases involving active respondents. This explains the topic's negative association with complaint success.

Not every topic can be easily understood. For instance, BTO\textsubscript{M} 57, represented by n-grams referencing Middle Eastern countries, indeed involved complainants from this region.\footnote{WIPO Case Nos.\ D2005-0309, D2008-0835, D2008-0895, D2009-0133, and D2015-0798.} Of these, 2 also involved Middle Eastern respondents. All 5 complaints were denied, but for differing reasons. In 3 cases the complainant failed to show bad faith because the domain had been registered \textit{before} the complainant's mark was established. Whether complaints from Middle Eastern parties are properly associated with these facts and with lower success rates is however unclear. Likewise, BTO\textsubscript{L} 1 is populated by n-grams tracking a typical portion in the ``Procedural History'' section which states that ``the Panel has submitted the Statement of Acceptance and Declaration of Impartiality and Independence, as required by the Center to ensure compliance with the Rules, paragraph 7''.\footnote{e.g.\ WIPO Case Nos.\ D2006-0874, D2006-1054, D2011-1122.} When this sentence occurs immediately before the next section header, ``Factual Background'', the topic's n-grams arise (after stopword removal). Why this correlates with better complainant outcomes is not clear. It may signal the lack of other procedural issues, such that panellists can move directly to the next section.\footnote{e.g.\ WIPO Case No.\ D2010-0593.}, but more qualitative evaluation is needed to ascertain this.

\subsection{ECHR Results}

Table \ref{tab:echr_res} presents results for LSA, BTO\textsubscript{M}, and BTO\textsubscript{L} regressions fit on ECHR case. As with the UDRP, LSA tends to produce higher model fits and the largest number of significant topics. This especially for Article 3, where 15 LSA topics are significant at ($\alpha=0.05$) compared to 3 BTO\textsubscript{M} and 4 BTO\textsubscript{L} topics. This is notable given that LegalBERT was finetuned on ECHR cases \cite{chalkidis-etal-2020-legal}. It is thus not surprising that BTO\textsubscript{L} again produces higher fit measures than BTO\textsubscript{M}, especially for Article 8 (AUROC=0.711 versus 0.638). However, both BTO models produce broadly similar numbers of selected and significant topics. 

\input{tables/echr_res}

\input{tables/echr_words}
Table \ref{tab:echr_words} presents representative n-grams for significant ECHR topics chosen based on smallest p-value and largest coefficient sizes. For Article 3 (prohibition torture or inhuman or degrading treatment), LSA 2 and BTO\textsubscript{L} 1 correctly discover and assign positive effects to what the ECtHR has described as ``a whole series of cases concerning allegations of disappearances in the Chechen Republic''.\footnote{HUDOC Case No.\ 001-140017.} Applicants were typically Chechen individuals whose close relatives were allegedly abducted by state military servicemen. Despite multiple complaints to and visits from the state's district prosecutor's office, the applicants hear nothing of their relatives for years. The ECtHR has ``found on many occasions'' that the distress caused by their relatives' disappearance and the state's indifference to their plight violates Article 3. The top cases for these topics all involved similar fact patterns\footnote{HUDOC Case Nos.\ 001-140017, 001-112097, 001-95882, 001-95457, 001-92119, 001-93121, 001-150311, 001-146390, and 001-70853.} BTO\textsubscript{L} 21 captures cases involving rejected asylum seekers who argued that they faced real risks of being subjected to treatment violating Article 3 if they were sent home. This allegedly because of their previous membership in military organisations that had clashed with their countries' current governments. ``December 2010'' is a significant n-gram because the United Nations High Commissioner for Refugees had issued updated eligibility guidelines for Afghan asylum seekers then. The top 5 cases were all complaints from ex-Afghan security service personnel. As the negative coefficient suggests, these claims were typically denied because, among other reasons, these guidelines did not include them in their risk profiles for rights violations.\footnote{HUDOC Case Nos.\ 001-113328, 001-57451, 001-60924, 001-146372, 001-69022.} Notably, there is a similar line of unsuccessful complaints involving failed asylum seekers who previously served in the Sri Lankan Tamil Tigers.\footnote{HUDOC Case Nos.\ 001-102949, 001-102955, 001-102947, 001-102957, 001-104956.} These were also picked up by LSA 12 (not tabulated in table \ref{tab:echr_words}), which is represented by n-grams including ``sri lanka'', ``ltte'', and ``colombo''. LSA 3 also identifies cases involving unsuccessful asylum seekers, but includes more varied claims from individuals originally from Somalia, Iraq, and Libya.\footnote{001-145018, 001-118339, 001-145789, 001-126027, 001-141949.} These complaints tended to fail because the court did not find a sufficiently real risk of treatment contrary to Article 3.

For Article 6 (right to fair trial), LSA 2 and 6 surface a collection of cases where Ukrainian individuals awarded compensation judgments against certain (often state-linked) companies were forced to wait for years before receiving due payment. They argued that the state bailiff had inordinately delayed enforcement proceedings. Decisions for such cases are worded very similarly, and typically reiterate how the ECtHR has ``already'' or ``frequently found'' violations in like cases.\footnote{HUDOC Case Nos.\ 001-91393, 001-71592, 001-93886, 001-75842, 001-78383, 001-78528, 001-78530, 001-75842, 001-78397, and 001-70357.} Notably, a Ukrainian government judicial enforcement reform effort acknowledges Article 6 as its motivation.\footnote{\url{https://www.kmu.gov.ua/en/reformi/verhovenstvo-prava-ta-borotba-z-korupciyeyu/reformuvannya-sistemi-vikonannya-sudovih-rishen}} BTO\textsubscript{L} 79 is diluted by markup n-grams like ``level0'' but nonetheless identifies several cases involving Cypriot individuals who had participated in a 1989 anti-Turkish demonstration in disputed territory arising out of the 1974 Turkish intervention in North Cyprus.\footnote{HUDOC Case Nos.\ 001-169203, 001-61582, 001-139903, 001-113876, 001-139995.} They were charged and convicted in the Turkish courts for entering Turkish territory without permission. Typically they argued that their Article 6 rights had been violated because the legal proceedings were generally in Turkish, not Greek which they understood. The ECtHR generally rejected these claims because the applicants had reasonable access to interpreters and other legal assistance.

For Article 8 (right to respect for family and private life, home, and correspondence), LSA 2 points to a line of cases filed against the Polish authorities by individuals detained in criminal remand over the authorities' standard practice of reading and re-sealing correspondence sent by these individuals to the courts and stamping the envelopes with a ``censored'' label. The ECtHR has noted how it has ``held on many occasions'' that the label forces the court to assume an interference with correspondence that, unless justified, violates Article 8.\footnote{HUDOC Case No.\ 001-93604 at paragraph 94.} 2 of the top 5 BTO\textsubscript{M} 70 cases have similar facts, but the topic also seems to cover other kinds of interferences to correspondence. Relatedly, the top 5 LSA 10 cases all involve complaints filed by Romani persons against the British government and its consistent refusal to grant them planning permission to develop land they owned into caravan sites. After the ECtHR found this to be a violation in 1996,\footnote{HUDOC Case No.\ 001-58076.} several similar and ultimately successful cases were raised in which the ECtHR expressly ``recalls that it has already examined'' such complaints and found violations.\footnote{HUDOC Case Nos.\ 001-59156, 001-59154, 001-59158, 001-59157.}

\section{Discussion}
\label{sec:discussion}

We proposed and evaluated an automated pipeline for discovering significant topics from cases by performing penalised regression and selective inference on features synthesized from decision texts using topic models. We show that significant topics discovered through this process capture relevant information on factual patterns correlated with case outcomes. On a large (by legal standards) dataset of UDRP cases, legal outcome models fit with decision text topics consistently produce higher fit scores compared with models fit without (table \ref{tab:udrp_res}). The LASSO also tends to select the topics as significant predictors over other structured case attributes of potential interest, such as judge identities. Coefficients and p-value estimates also change noticeably. This holds across several regression settings and topic modeling approaches. Using a canonical dataset of ECHR cases, we show that the method generalises relatively easily, without the need for additional feature engineering or pre-processing. Only structured outcome information and unstructured decision texts are required, though additional variables can be added at the regression step. Running similar procedures on the existing dataset, albeit with corpus-tailored hyperparameters (such as topic number and $\lambda$), yields significant topics consistent with ECHR case law.

Our experiments show that LSA is a useful, even if dated, codebook for decision texts. Across all experiment settings, LSA produced higher fit scores and a higher number of significant topics than both BTO models. LSA is also computationally cheaper. This may appear counter-intuitive since BTO is a significantly more sophisticated model which exploits recent advances like word embeddings and language modelling. As noted by \cite{soh_legal_2019} in the context of legal topic classification, the length of legal decision texts may offer one explanation for LSA out-performing newer approaches in the legal domain. BTO's superiority over traditional topic models has mainly been demonstrated on shorter texts like tweets and news articles \cite{grootendorst2022bertopic, degroot2022experiments}. For longer documents, our present approach of reconstituting document-level topics by normalising chunk-level topic counts is, while standard in the literature \cite{silveira_et_al_2021}, unlikely to be the optimal way to deploy BTO. Using LMs with larger context windows than those tested here could significantly improve BTO's performance. Thus, we do not suggest that LSA is necessarily better-suited for this task. Further, since each topic model yields different topics which may provide different insights on the cases, there is no clear metric for ``better'' in this context. Further, BTO's lower fit scores may be a methodological artefact since we did not conduct hyperparameter optimisation, but chose similar parameters across all topic models to establish a baseline comparison. A fully-optimised BTO model may outperform a fully-optimised LSA. Hyperparameter tuning was not done because, unlike typical machine learning settings, our task focuses on explanation rather than classification and does not yield any clear performance metric (e.g.\ F1 score) for evaluating a grid search. BTO's performance here should be interpreted in this light.

More importantly, qualitative evaluation of significant topic n-grams demonstrates that the discovered topics rest on sound and interpretable legal bases. For UDRP cases, the topics shed correlative light on how administratively deficient complaints are less likely to win, how famous trademark owners can be associated with higher success rates, and how respondents who actively defend their domains are less likely to lose. For Arts 3, 6, and 8 ECHR, the topics identify archetypal cases involving abducted Chechen relatives, Afghan asylum seekers, Ukrainian judgment enforcers, Cypriot demonstrators, Polish detainees, and Romani land owners. These are correctly associated with their usual case outcomes. Essentially, unique case features prompt judges into writing decisions with a higher preponderance of correspondingly unique n-grams, producing signals which the topic models are capable of detecting. As legal decisions are written with close reference to case facts and relevant laws, and judges would generally not write about irrelevant matters and non-issues, we theorise that the decision text generation function accords with the standard topic modelling assumption that texts are generated by sampling n-grams from latent topics \cite{blei_ng_jordan_2003}. 

To be sure, not every discovered topic made sense. This may point to limitations in our evaluation process, since we only sampled the top 5 cases associated with the most significant topics. We may also have been unable to detect known patterns that the topics were in fact referencing. Further, there is no reason why each topic should capture exactly one predictor. Certain topics may have had n-gram distributions amalgamating several archetypal case features. Individually insignificant topics could have been jointly significant with others. Future work could examine this further by modeling interaction terms and conducting joint tests, though this may make interpreting the topics and coefficients more challenging for evaluators. Besides human limitations, the automated process is also imperfect. It can produce false positives (e.g.\ significant topics which do not actually capture legal predictors; high document-topic weights for a case not actually on topic) and false negatives (not attaching significance to topics which do; not synthesising a related topic to begin with).

At a more abstract level, the four pipeline steps can be understood as a series of dimensionality reduction steps, starting with a large decision corpus, and resulting in numerical associations between decisions and topics (document-topic weights), topics and outcomes (regression coefficients), and topics and words (n-gram distributions). These mappings can be analysed transitively to assist with the discovery, extraction, and analysis tasks identified in part \ref{meth:defn}. To illustrate, after observing UDRP LSA topic 19 (table \ref{tab:udrp_words}), researchers could create and extract observations for an indicator variable for whether the complainant owned a famous mark. Decision-topic weights produced by the method could guide the extraction process. After several such variables are extracted, a regression (not necessarily the LASSO) could then be run on the reduced dataset. Notably, given the increasing popularity of large language models, the pipeline's ability to reduce large legal corpora to smaller components could prove useful in fitting legal texts into limited context windows. 

\section{Conclusion}

We proposed and assessed an automated method for discovering significant topics given only decision texts and case outcomes, building on prior work examining how topic models can be used to predict and explain case outcomes \cite{aletras_2016, falakmasir_2017, salaun_why_2022}. The task of legal topic discovery was formally defined and distinguished from related identification and analysis tasks. We developed and demonstrated pre-processing, topic modelling, regression and inference steps tailored to this task and its legal context. The method shows promise in its ability to discover archetypal case features and patterns consistent with the jurisprudence of the UDRP and ECHR datasets tested, and could generalise to other areas. It is however not perfect, and should be applied bearing the possibility of false positives and negatives in mind. There are two extensions we hope to pursue in future work. First, to conduct more rigorous experiments and hyper-parameter search with BERTopic and its variations. Notably, BERTopic is a modular framework involving six steps that accept several different algorithms and (optional) parameters. Second, a more robust yet ideally less manual method for evaluating and interpreting topics could be developed.

\singlespacing
\bibliographystyle{plain}
\setlength{\bibsep}{0pt}
\bibliography{biblio.bib}

\end{document}

%% file: tables/task_table.tex
\begin{table}[ht]
\centering
\smalltablefont
        \begin{tabularx}{\textwidth}{p{5cm}p{5cm}X}
        \toprule
        Given Inputs & Goal & Task Label(s) \\ \midrule
        $\hat{X}, \hat{W}$ & Estimate weights & Analysis \\
        $D$, candidate $\hat{X}, \hat{W}$s known & Extract observations of $\hat{X}, \hat{W}$ & Identification/Extraction \\
        $D$, candidate $\hat{X}, \hat{W}$s unknown & Discover candidate $\hat{X}, \hat{W}$s & Discovery \\
        \bottomrule
        \end{tabularx}
    \flushleft{
    \smalltablenotefont
    \caption{} \label{tab:task_names} 
        Tasks involving legal predictors. This work focuses on the discovery task.
    }
\end{table}

%% file: tables/udrp_stats.tex
\begin{table}
    \centering
    \smalltablefont
        \begin{tabularx}{\textwidth}{@{}Xrrr@{}}
        \toprule
                             & Complainant Won  & Complainant Lost & Overall \\ 
        \midrule
        No.\ comp'ts     & 1.108            & 1.095         & 1.106     \\
                             & (0.393)          & (0.392)       & (0.393) \\
        No.\ resp'ts      & 1.094            & 1.077         & 1.092     \\
                             & (1.301)           & (0.349)    & (1.232) \\
        No.\ domain names     & 2.009            & 1.407         & 1.942     \\
                             & (14.749)          & (2.188)    & (13.921) \\
        Raw word count       & 2,812.439        & 3,560.197     & 2,895.986 \\
                             &(1,309.640)       & (1,680.681)   & (1,376.408) \\
        Processed word count & 1,293.598        & 1,630.689     & 1,331.261 \\
                             & (600.045)         & (772.365)     & (630.659) \\
        GTLD cases (\%)           & 94.32       & 94.94         & 94.39 \\
        Three-member Panel Cases (\%) & 3.21        & 24.50         & 5.59  \\
        Comp't Won (\%)      & 100.00      & 0.00          & 88.83 \\
        N             & 20122         & 2531          & 22653 \\
        \bottomrule
        \end{tabularx}
        \flushleft{
        \caption{} \label{tab:udrp_stats}
        \smalltablenotefont
        UDRP summary statistics by outcome. Mean values presented. Standard deviations in brackets. Raw word count includes all tokens in the text after removing only the ``Decision'' section. Processed word counts includes only tokens remaining after lower-casing, stopword removal, and lemmatisation were further applied.
        }
\end{table}

%% file: tables/echr_stats.tex
\begin{table}[ht]
\centering
\smalltablefont
    \begin{tabularx}{\textwidth}{lrrrrrr}
    \toprule
     Article Contested: & \multicolumn{2}{c}{Art 3} & \multicolumn{2}{c}{Art 6} & \multicolumn{2}{c}{Art 8} \\
     \cmidrule(lr){2-3}\cmidrule(lr){4-5}\cmidrule(lr){6-7}
     & Violation & No Violation & Violation & No Violation & Violation & No Violation \\
    \midrule
    Raw Word Count & 4313.676 & 5135.169 & 1768.104 & 3894.557 & 4063.601 & 4909.45 \\
     & (3579.273) & (4046.314) & (2054.486) & (2894.177) & (4003.777) & (2897.937) \\
    Proc'd Word Count & 2020.648 & 2424.053 & 821.394 & 1800.873 & 1891.228 & 2277.022 \\
     & (1680.586) & (1916.545) & (943.129) & (1344.95) & (1893.032) & (1353.649) \\
    N\ & 284 & 284 & 454 & 449 & 228 & 229 \\
    \bottomrule
    \end{tabularx}
\flushleft{
    \caption{}
    \label{tab:echr_stats}
    \smalltablenotefont
    Summary statistics for the ECHR dataset. Mean values presented with standard deviations in brackets. Notice that \cite{} had balanced the dataset by random under-sampling.
}
\end{table}

%% file: tables/udrp_res.tex
\renewcommand{\arraystretch}{0.8}
\begin{table}
\centering
\smalltablefont
\begin{tabularx}{\textwidth}{@{}Xrrrrrrr@{}}
\toprule
Topic Model: & \multicolumn{3}{c}{None} & \multicolumn{1}{c}{OHE} & \multicolumn{3}{c}{LSA} \\
\cmidrule(lr){2-4}\cmidrule(lr){5-5}\cmidrule(l){6-8}
Y: Complaint Success (binary) & (1) & (2) & (3) & (4) & (5) & (6) & (7) \\
\midrule
\textbf{Panellists} & & & & & & &\\
Median Coef & -0.0611 & -0.0308 & -0.0306 & -0.037 & -0.0397 & -0.0222 & -0.0241 \\
Median PV & 0.0677 & 0.236 & 0.179 & 0.401 & 0.336 & 0.376 & 0.409 \\
No.\ Sig.\ ($\alpha=0.05$) & 53 & 50 & 49 & 19 & 9 & 13 & 8 \\
No.\ Sig.\ ($\alpha=0.01$) & 41 & 22 & 27 & 18 & 5 & 6 & 0 \\
No.\ Sig.\ ($\alpha=0.001$) & 12 & 8 & 19 & 1 & 1 & 1 & 0 \\
No. Selected & 117 & 159 & 151 & 117 & 68 & 101 & 102 \\
Total No. & 462 & 501 & 510 & 510 & 462 & 501 & 510 \\
\addlinespace[2ex]
Panel Size &  & -0.154 & -0.146 & -0.0755 &  & -0.138 & -0.128* \\
 &  & (0.024) & (0.023) & (0.023) &  & (0.0332) & (0.0236) \\
\textbf{Topics} & & & & & & &\\
Median Coef &  &  &  & 0.0374 & 0.0382 & -0.0352 & -0.0363 \\
Median PV &  &  &  & 0.269 & 0.389 & 0.183 & 0.167 \\
No.\ Sig.\ ($\alpha=0.05$) &  &  &  & 7 & 24 & 31 & 32 \\
No.\ Sig.\ ($\alpha=0.01$) &  &  &  & 2 & 16 & 20 & 14 \\
No.\ Sig.\ ($\alpha=0.001$) &  &  &  & 1 & 9 & 8 & 6 \\
No. Selected: &  &  &  & 37 & 92 & 103 & 105 \\
\midrule
Setting$^a$ & A & B & C & C & A & B & C \\
Median Dev.\ Ratio & 0.148 & 0.227 & 0.224 & 0.328 & 0.389 & 0.443 & 0.431 \\
AUROC & 0.742 & 0.824 & 0.822 & 0.874 & 0.903 & 0.914 & 0.914 \\
\midrule
\addlinespace[1ex]
 & \multicolumn{3}{c}{BTO\textsubscript{M}} & \multicolumn{3}{c}{BTO\textsubscript{L}} & \multicolumn{1}{c}{Noise} \\
\cmidrule(lr){2-4}\cmidrule(lr){5-7}\cmidrule(l){8-8}
  & (8) & (9) & (10) & (11) & (12) & (13) & (14) \\
\midrule
\textbf{Panellists}\\
Median Coef & -0.0534 & -0.0288 & -0.0298 & -0.0615 & -0.0328 & -0.0363 & -0.0293 \\
Median PV & 0.12 & 0.24 & 0.265 & 0.219 & 0.297 & 0.236 & 0.166 \\
No.\ Sig.\ ($\alpha=0.05$) & 39 & 40 & 47 & 25 & 42 & 8 & 52 \\
No.\ Sig.\ ($\alpha=0.01$) & 32 & 30 & 26 & 17 & 13 & 5 & 33 \\
No.\ Sig.\ ($\alpha=0.001$) & 22 & 16 & 9 & 9 & 2 & 4 & 24 \\
No. Selected & 97 & 141 & 150 & 95 & 143 & 132 & 151 \\
Total No.\ & 462 & 501 & 510 & 462 & 501 & 510 & 510 \\
\addlinespace[2ex]
Panel Size &  & -0.185 & -0.167 &  & -0.125 & -0.113 & -0.143 \\
 &  & (0.0239) & (0.0233) &  & (0.0245) & (0.0234) & (0.023) \\
\textbf{Topics} & & & & & & &\\
Median Coef & -0.0267 & 0.027 & -0.0114 & 0.032 & 0.0444 & 0.0295 & 0.0352 \\
Median PV & 0.0854 & 0.351 & 0.317 & 0.206 & 0.654 & 0.348 & 0.58 \\
No.\ Sig.\ ($\alpha=0.05$) & 27 & 11 & 11 & 20 & 17 & 15 & 1 \\
No.\ Sig.\ ($\alpha=0.01$) & 15 & 4 & 4 & 10 & 10 & 9 & 0 \\
No.\ Sig.\ ($\alpha=0.001$) & 6 & 0 & 0 & 4 & 4 & 6 & 0 \\
No. Selected: & 69 & 97 & 96 & 67 & 79 & 85 & 30 \\
\midrule
Setting$^a$ & A & B & C & A & B & C & C \\
Median Dev.\ Ratio & 0.222 & 0.274 & 0.274 & 0.215 & 0.29 & 0.295 & 0.237 \\
AUROC & 0.789 & 0.836 & 0.838 & 0.798 & 0.846 & 0.849 & 0.825 \\
\bottomrule
\end{tabularx}
\flushleft{
\caption{}
\label{tab:udrp_res}
\smalltablenotefont
LASSO logit regression results for UDRP cases. Given the number of panellists and topics input we report medians and counts instead of individual estimates. Coefficients are direct estimates from \texttt{glmnet} and should not be interpreted cardinally. Standard errors in brackets.\\
$^a$ Data partition and controls used. Setting A includes 20150 1-panelist GTLD cases, excludes $controls_i$ (see part \ref{sec:data:udrp}), and sets $\lambda=68.454$. B includes all 21383 GTLD cases, includes $controls_i$, and sets $\lambda=64.068$. C includes 22653 GTLD/CCTLD cases, includes $controls_i$, and sets $\lambda=61.784$. 
}

\end{table}
\renewcommand{\arraystretch}{1} 

%% file: tables/udrp_words.tex
\begin{table}
\smalltablefont
\begin{tabularx}{\textwidth}{lllX}
\toprule
Model & Topic & Coefficient & Representative N-grams\\
\midrule
\multirow[t]{5}{*}{
LSA} & 19 & 0.6069*** & trade mark, lego, trade, famous, world, wipo case, amendment complaint, amendment, brand \\
 & 28 & 0.1724*** & asserts, trade mark, complainant asserts, argues, complainant argues, alleges, complainant alleges, alleges respondent, trade \\
 & 3 & -0.3724*** & contact information, registrant contact information, registrant contact, information, amended complaint, amended, amendment, amendment complaint, disclosed \\
 & 17 & -0.4683*** & trade mark, trade, amendment, deficient, administratively deficient, administratively, complaint administratively, complaint administratively deficient, amendment complaint \\
 & 2 & -0.7685*** & administrative, copy, e-mail, received, icann, notification, administrative panel, registrar domain, registrar domain name \\
\midrule
\multirow[t]{5}{*}{
BTO\textsubscript{M}} & 5 & 0.1681** & armani, ikea, boss, bmw, reg, classes, elite, hugo, hugo boss, international trademark \\
 & 6 & 0.1517** & pharmaceutical, sanofi, pfizer, sanofiaventis, aventis, 100 countries, prescription, drug, treatment, weight \\
 & 11 & 0.1132** & chase, cme, barclays, financial services, bank, nasdaq, financial, insurance, investment, banking \\
 & 199 & -0.044* & videos, sports, action, complaint exhibit, jeff, complaint exhibit respondent, january 31 2000, complainant action, skiing, omit \\
 & 57 & -0.0747** & qatar, al, emirates, arabic, discover, project, uae, abu, brothers, dubai \\
\midrule
\multirow[t]{5}{*}{
BTO\textsubscript{L}} & 1 & 0.2028*** & rules paragraph factual, paragraph factual background, paragraph factual, factual background complainant, background complainant, panel submitted statement, ensure compliance rules, impartiality independence required, independence required, ensure compliance \\
 & 66 & 0.1911*** & remedy transfer, support case, registered subsequently used, registered subsequently, elements complainant, subsequently used, complainant support, respondent transferred, remedy, policy domain \\
 & 21 & 0.1645*** & publicdomainregistrycom, dba publicdomainregistrycom, pvt dba, directi internet solutions, directi internet, internet solutions, directi, internet solutions pvt, solutions pvt, pvt \\
 & 204 & -0.0672*** & oy, page displayed, banners, english version, portal, illegally, marks owned, domain names redirect, names redirect, alex \\
 & 69 & -0.0754*** & domain hijacking, reverse domain hijacking, hijacking, reverse domain, respondent requests, reverse, finding reverse, respondent requests panel, finding reverse domain, domain hijacking complainant \\
\cline{1-4}
\bottomrule{}
\end{tabularx}
\flushleft{
    \caption{}
    \label{tab:udrp_words}
    UDRP topics with smallest p-values across Setting C regressions \ref{tab:udrp_res}.7, \ref{tab:udrp_res}.10, and \ref{tab:udrp_res}.13. Coefficients are scaled estimates from the LASSO and should only be interpreted ordinally within the same regression. Topics are synthesized from masked decision texts that exclude ``Discussion and Findings'' and later sections and not be interpreting as capturing what the panels found. $\text{*}  p < 0.05, \text{**}  p < 0.01, \text{***}  p < 0.001$
}
\end{table}

%% file: tables/echr_res.tex
\begin{table}
\centering
\smalltablefont
\begin{tabularx}{\textwidth}{@{}Xrrrrrrrrr}
\toprule
Y: Violation Found (binary) & \multicolumn{3}{c}{Article 3} & \multicolumn{3}{c}{Article 6} & \multicolumn{3}{c}{Article 8} \\
\cmidrule(lr){2-4}\cmidrule(lr){5-7}\cmidrule(l){8-10}
Topic Model: & LSA & BTO\textsubscript{M} & BTO\textsubscript{L} & LSA & BTO\textsubscript{M} & BTO\textsubscript{L} & LSA & BTO\textsubscript{M} & BTO\textsubscript{L} \\
\midrule
Median Coef & -0.164 & -0.161 & -0.144 & 0.153 & -0.185 & -0.119 & -0.005 & 0.041 & -0.129 \\
Median PV & 0.02 & 0.177 & 0.088 & 0.521 & 0.171 & 0.324 & 0.211 & 0.257 & 0.451 \\
No.\ Sig.\ ($\alpha=0.05$) & 15 & 3 & 4 & 5 & 2 & 3 & 3 & 1 & 0 \\
No.\ Sig.\ ($\alpha=0.01$) & 13 & 1 & 4 & 3 & 1 & 1 & 2 & 0 & 0 \\
No.\ Sig.\ ($\alpha=0.001$) & 10 & 0 & 2 & 1 & 1 & 1 & 1 & 0 & 0 \\
No.\ Selected & 28 & 13 & 9 & 17 & 5 & 21 & 10 & 4 & 16 \\
\midrule
N & 568 & 568 & 568 & 903 & 916 & 916 & 457 & 458 & 458 \\
$\lambda$ & 15.279 & 24.603 & 23.029 & 22.087 & 42.502 & 22.136 & 20.09 & 25.035 & 17.599 \\
Median Dev.\ Ratio & 0.508 & 0.327 & 0.326 & 0.384 & 0.223 & 0.266 & 0.344 & 0.253 & 0.286 \\
AUROC & 0.89 & 0.73 & 0.739 & 0.855 & 0.746 & 0.784 & 0.763 & 0.638 & 0.711 \\
\bottomrule
\end{tabularx}
\flushleft{
\caption{}
\label{tab:echr_res}
\smalltablenotefont
LASSO logit results for ECHR cases. $\lambda$s are separately derived per model following \cite{negahban_2009}. 
}
\end{table}

%% file: tables/echr_words.tex
\begin{table}
\smalltablefont
\begin{tabularx}{\textwidth}{@{}lllX@{}}
\toprule
Art & Topic & Coefficient & Representative n-grams (by descending weight)\\
\midrule
\multirow[t]{6}{*}{3} & LSA 2 & 0.561*** & servicemen, abduction, district prosecutor office, district prosecutor, prosecutor office, chechen, chechnya, military, prosecutor \\
 & BTO\textsubscript{L}  1 & 0.4994*** & abduction, military, prosecutors office, men, relatives, prosecutors, criminal case, identify, forwarded, armed \\
 & BTO\textsubscript{M}  7 & 0.4069* & athe, bthe, governments, account, version, 1the, events, applicants detention, cthe, 2the \\
 & BTO\textsubscript{L}  21 & -0.4696*** & unhcr, december 2010, groups, sri, violations, 1951, lanka, sri lanka, international, refugees \\
 & LSA 10 & -0.7249*** & regional, regional court, burned, villages, villagers, village, houses, pkk, appeal \\
 & LSA 3 & -1.3226*** & asylum, country, refugee, board, unhcr, kingdom, forces, united, members \\
\midrule
\multirow[t]{6}{*}{6} & LSA 2 & 1.5781*** & court, applicant, enforcement, russian, bailiffs, ukraine, ukrainian, uah, enforcement proceedings \\
 & LSA 6 & 0.3544* & ukraine, uah, ukrainian, bailiffs, bailiffs service, debtor, turkish, v., law relevant \\
 & LSA 11 & 0.2906** & applicants, tenant, land, italian, cell, property, detention, possession, treatment \\
 & BTO\textsubscript{L}  79 & -0.2929* & greek, territory, entry, called, witnesses, level0, level0 arabic, seq level0, seq level0 arabic, arabic \\
 & BTO\textsubscript{L}  0 & -0.3873*** & 3the applicant, alleged, 3the, detention, access, complained, custody, statements, particular, torture \\
 & BTO\textsubscript{M}  0 & -0.5878*** & imprisonment, years, expert, village, tax, damage, social, seq, medical, regional court \\
\midrule
\multirow[t]{4}{*}{8} & LSA 2 & 0.822*** & detention, prosecutor, criminal, remand, applicant detention, criminal proceedings, regional, regional court, applicant \\
 & LSA 10 & 0.3846** & hague convention, russian, poland, gypsies,$^\alpha$ polish, gypsy,$^\alpha$ hague, retention, sites \\
 & BTO\textsubscript{M}  70 & 0.3753* & correspondence, prohibition, mean, content, shown, february 2003, ask, world, avoid, 9the \\
 & LSA 7 & -0.3084* & prison, detention, expulsion, aliens, residence permit, detained, visits, asylum, cell \\
\bottomrule
\end{tabularx}
\flushleft{
    \caption{}
    \label{tab:echr_words}
    \smalltablenotefont
    Significant ECHR topics across all Articles and topic models tested. We select topics to report here by first identifying the 5 lowest p-value topics within each regression, and then choosing topics with the 3 most positive and negative effects across all 3 regressions within each Article. Article 8 has only 4 significant topics in total. $\text{*}  p < 0.05, \text{**}  p < 0.01, \text{***}  p < 0.001$ \\
    $^\alpha$ We note that this token is derogatory and have preferred the term Romani in the main article. With apologies to the Romani people, it was retained here to reflect tokens actually used in the corpus. 
}
\end{table}